\documentclass[10pt,twocolumn,letterpaper]{article}
\usepackage[T1]{fontenc}
\usepackage{paperlayout}
\usepackage{amsmath}
\definecolor{linkblue}{rgb}{0.21,0.49,0.74}
\usepackage[pagebackref,breaklinks,colorlinks,allcolors=linkblue]{hyperref}
\usepackage[capitalize]{cleveref}
\crefname{section}{Sec.}{Secs.}
\Crefname{section}{Section}{Sections}
\Crefname{table}{Table}{Tables}
\crefname{table}{Tab.}{Tabs.}
\usepackage{afterpage}
\title{From VLM Planners to Robotic Agents: Self-Evolving Harnesses for History-Dependent Manipulation}
\author{\small\bfseries Tingwen Du$^{1}$\enspace Haoyu Guo$^{2,\dagger}$\enspace Hanxue Zhang$^{3}$\enspace Xuanxu Wang$^{3}$\enspace Junting Dong$^{2}$\enspace Chao Gao$^{2}$\\
\small\bfseries Mingfei Han$^{4}$\enspace Tao Lu$^{2}$\enspace Mulin Yu$^{2}$\enspace Chunhua Shen$^{5}$\enspace Xiaojun Chang$^{1,\ddagger}$\enspace Zhihui Li$^{1}$\\[0.5em]
\normalfont\small $^{1}$University of Science and Technology of China\\
\normalfont\small $^{2}$Shanghai Artificial Intelligence Laboratory\quad $^{3}$Shanghai Jiao Tong University\\
\normalfont\small $^{4}$Mohamed bin Zayed University of Artificial Intelligence\quad $^{5}$Zhejiang University\\[0.4em]
\normalfont\small $^{\dagger}$Project Leader\quad $^{\ddagger}$Corresponding author}

\begin{document}
\raggedbottom
\maketitle

\begin{abstract}
Vision-language-action (VLA) models achieve strong performance
across embodied manipulation tasks by unifying visual understanding,
language conditioning, and action generation.\space However, long-horizon tasks can involve non-Markovian dependencies: successful
decisions may require retaining task progress and past interaction outcomes,
while partial observability and execution errors compound over time.\space Existing work has begun to address these challenges by augmenting
foundation-model policies with memory and by introducing hierarchical
architectures that separate high-level planning from low-level execution.\space {We organize the high-level VLM as a robotic agent through an editable
harness to maintain task context and improve planning through execution
experience.\space
We instantiate this approach as {SERA (Self-Evolving Robotic Agent)}, with a
fixed VLA executor providing low-level control.}\space
The harness maintains and updates task context using observations and execution
feedback to guide subsequent planning.\space {In simulation, self-evolution uses environment rollouts to propose and evaluate}
harness revisions, with the selected configuration fixed before held-out
testing.\space On RoboMME, {SERA} achieves 51.38\% overall success with autonomous selection,
compared with 44.38\% without self-evolution.\space
{We also evaluate history-dependent real robot tasks with
human-reviewed harness corrections.}\space Our exploration suggests that agentic planning is a promising direction for
history-dependent robotic manipulation.\space

\end{abstract}

\begingroup
\suppressfloats[t]
\begin{figure}[t]
\centering
\makebox[\linewidth][c]{\includegraphics[width=1.05\linewidth]{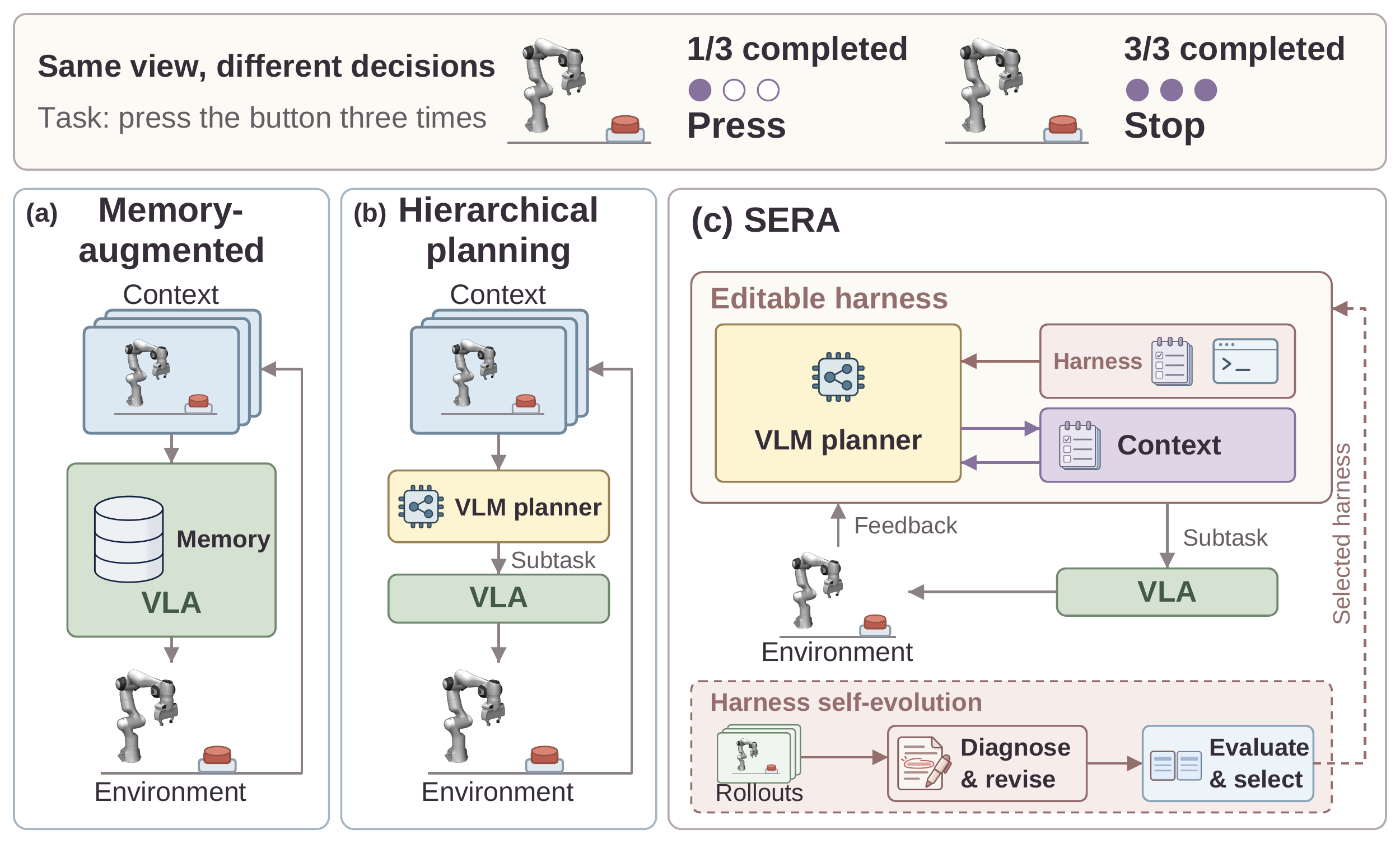}}
\caption{{\textbf{From planner to agent for history-dependent manipulation.}}
{Top: Similar observations can require different actions depending
on task progress. (a) Memory-augmented VLAs retain historical information.
(b) Hierarchical planners issue subtasks to low-level executors.
(c) SERA maintains task context through an editable harness, revised using
rollout feedback. The dashed arrow marks the selected harness, fixed before
\mbox{testing}.}}
\label{fig:motivation}
\end{figure}

\section{Introduction}
\label{sec:introduction}

Vision-language-action (VLA) models connect visual observations and language
instructions to robot actions, providing a foundation for general-purpose
manipulation~\citep{brohan2023rt2,kim2024openvla,black2024pi0}.
Long-horizon tasks, however, can require information that the current observation
does not reveal~\citep{koo2025hamlet,dai2026robomme}. For example, when a robot
must repeat an operation a specified number of times, similar images may call
for different decisions depending on how many repetitions have actually
succeeded. An earlier observation may also contain a cue that is no longer
visible. Such history-dependent decisions require preserving task-relevant
information and tracking execution outcomes over successive interactions.

Existing research provides several foundations for these capabilities.
Memory-augmented policies equip VLA models with mechanisms for retaining and
using historical information, including compact temporal representations
\citep{koo2025hamlet,jang2025contextvla,zhu2026weavela}, memory retrieval, and
recurrent state updates~\citep{shi2025memoryvla,hu2026echo,li2026rememvla}.
Hierarchical architectures introduce a high-level planning layer that
issues intermediate decisions to low-level executors
\citep{ahn2022saycan,belkhale2024rth,duan2025robopara,sridhar2025memer} and can
incorporate environment feedback~\citep{huang2022innermonologue}. Agentic systems further
organize reasoning, actions, and feedback into iterative decision processes
\citep{yao2023react,shinn2023reflexion,wang2024voyager}. Recent embodied harnesses
explore external task management~\citep{zhang2026harnessvla,gu2026harnesswam}
and experience-driven adaptation
\citep{chen2026eta,wang2026shaper,ding2026zetta}.
Together, these developments motivate studying how the high-level planner in a
hierarchical architecture can manage task context and make use of execution
experience.

Our focus is on two connected responsibilities of this planning process.
Within an episode, issuing a command does not establish that the intended
operation has succeeded. In the repetition example, updating progress requires
interpreting subsequent observations and feedback, while retaining information
needed for later decisions. Access to history therefore needs to be accompanied
by an explicit process for maintaining decision-relevant task context.
Across attempts, an execution record must be translated into a reusable change
to influence future planning. This raises a question: how can we support both
ongoing context management and improvements to the planning process from
execution experience? We explore an editable harness around the high-level VLM
as a common mechanism for these responsibilities. The harness organizes context
and feedback during execution, while providing an explicit object for revision
during self-evolution. This also offers a way to investigate experience reuse
when additional demonstrations and manual annotations are costly.
Figure~\ref{fig:motivation} illustrates this design direction.

We present {SERA (Self-Evolving Robotic Agent)}, which organizes a VLM as an
agent within an editable harness around a fixed VLA executor. {During task execution, the harness organizes a workflow in which
the VLM interprets recent observations, updates retained task context, and
selects an executor command within the same planning call. This carries
information from earlier interactions into subsequent decisions.}
{However, a predefined harness may not anticipate all task requirements,
motivating execution-driven self-evolution to identify and repair its limitations.}
Analysis and revision agents autonomously diagnose failures from execution
traces and propose {candidate} repairs to planning guidance, assessment
criteria, and control rules.
{{In simulation, a verifier evaluates these candidates through environment rollouts,}
providing feedback for further revision and harness selection.}
{This process revises the harness without manually constructing additional
demonstration datasets for supervised fine-tuning of the high-level VLM planner.}
The selected harness is fixed before held-out testing, while task context
continues to be updated within each episode.

We evaluate {SERA} on RoboMME, a benchmark for history-dependent robotic
manipulation~\citep{dai2026robomme}. Autonomous {SERA} achieves 51.38\% overall
success, compared with 44.38\% without self-evolution. {We also evaluate history-dependent manipulation on a real robot.} Our contributions are:

\begin{itemize}
    \item An agentic planning framework that organizes a high-level VLM within an
    editable harness for sustained task-context management over long horizons,
    tracking task progress and execution feedback to guide a fixed VLA executor.
    \item An execution-driven self-evolution procedure that autonomously proposes
    and evaluates revisions to planning guidance, assessment criteria, and
    control rules from rollout feedback, without constructing an SFT dataset.
    \item {SERA achieves 51.38\% overall success on RoboMME,
    with self-evolution improving success by 7.00 percentage points.
    On a real robot, SERA achieves 85.4\% overall success
    on history-dependent manipulation tasks.}
\end{itemize}

\endgroup
\section{Related Work}
\label{sec:related-work}

\subsection{Memory-Augmented Policies}

{Generalist robot policies learn language-conditioned manipulation from diverse
data~\citep{brohan2022rt1,openx2023rtx,octo2024octo}.}
{VLA models connect pretrained vision-language representations with action
learning} (\citealp{brohan2023rt2,kim2024openvla,black2024pi0};
{\citealp{nvidia2025groot,physicalintelligence2025pi05}}).
Related embodied models, including PaLM-E, incorporate visual and continuous
state inputs into language models for embodied reasoning and planning
\citep{driess2023palme}. These developments provide a foundation for
language-conditioned manipulation, while history-dependent tasks additionally
require information beyond the current observation. Recent work has also
emphasized the importance of historical information in manipulation, with
benchmarks such as RoboMME explicitly evaluating history-dependent
behavior~\citep{dai2026robomme}.

Memory-augmented policies incorporate such information through different
representations and update mechanisms~{\citep{fang2025sam2act}}. HAMLET aggregates compact moment tokens
across timesteps to condition action prediction~\citep{koo2025hamlet}.
MemoryVLA retrieves perceptual and cognitive information from a memory bank
and fuses it with current representations~\citep{shi2025memoryvla}, while
ContextVLA compresses past observations into a context token for efficient
multi-frame conditioning~\citep{jang2025contextvla}. Other designs address
cross-subtask information transfer through event-triggered latent memory
in WeaveLA~\citep{zhu2026weavela}, hierarchical experience retrieval in
ECHO~\citep{hu2026echo}, and temporal aggregation through frame- and
chunk-level recurrent queries in ReMem-VLA~\citep{li2026rememvla}.
These methods integrate historical information into action prediction through
compact temporal representations, memory retrieval, or recurrent updates.

\subsection{Hierarchical Robotic Planning}

Hierarchical approaches connect high-level reasoning to executable robot
behavior through intermediate decisions~{\citep{rana2023sayplan,shi2025hirobot,zhang2026lingbotva2}}. SayCan combines language-model
planning with affordance estimates for pretrained skills
\citep{ahn2022saycan}, and Inner Monologue incorporates environment feedback
into subsequent planning~\citep{huang2022innermonologue}. Code as Policies
expresses robot behavior as programs invoking perception and control APIs
\citep{liang2023codeaspolicies}. Other interfaces include the spatial value
maps of VoxPoser, the relational keypoint constraints of ReKep, and the
language motions of RT-H
\citep{huang2023voxposer,huang2024rekep,belkhale2024rth}.
RoboPARA further uses dependency graphs and resource constraints to organize
parallel dual-arm execution~\citep{duan2025robopara}.

Memory and hierarchical planning can also be combined. MemER trains a
high-level VLM to nominate historical keyframes and predict language subtasks;
selected keyframes inform subsequent decisions, while a low-level VLA executes
the subtasks~\citep{sridhar2025memer}. Goal2Skill combines structured task
memory with subtask verification and replanning in a VLM--VLA hierarchy
\citep{liu2026goal2skill}.

\subsection{Agentic Embodied Systems and Self-Evolution}

{Research on language models and agents studies reasoning, interaction, and
feedback-driven refinement~\citep{madaan2023selfrefine}.} ReAct interleaves reasoning with actions
and observations~\citep{yao2023react}; Reflexion stores textual feedback for
subsequent attempts~\citep{shinn2023reflexion}; and Voyager accumulates
reusable executable skills through interaction in Minecraft
\citep{wang2024voyager}. {Related methods improve prompts or agent code using execution
feedback~\citep{agrawal2025gepa,zhang2025dgm}.} These works provide precedents for agent loops and
experience reuse, although their evaluation settings differ from robotic
manipulation.

Recent embodied systems place these ideas around learned
executors~{\citep{ahn2024autort}}.
Harness VLA combines a frozen VLA primitive with analytic primitives and
task-specific and global memories~\citep{zhang2026harnessvla}.
HarnessWAM maintains evidence-grounded task state and a task graph, and
compiles plans into capability-constrained WAM skills
\citep{gu2026harnesswam}. RoboMemory combines spatial, temporal, episodic,
and semantic memories with a planner--critic loop~\citep{lei2025robomemory}.
HELM couples episodic memory with a learned state verifier and recovery control
around a frozen VLA~\citep{zeng2026helm}. {Cortex combines task-context memory
and skill constraints in a deployment harness~\citep{peng2026cortex}.}

Execution-driven harness adaptation is also an emerging research direction.
ETA studies experience-derived strategy and skill updates subject to runtime
checks before promotion~\citep{chen2026eta}.
SHAPER evolves textual skills and a context-code harness using rollout
feedback and validation while keeping its planner and executor frozen
\citep{wang2026shaper}. Zetta revises runtime critics, recovery playbooks,
and executable tools from failed rollouts~\citep{ding2026zetta}.
Within this line of work, {SERA} explores an editable harness for high-level
planning in history-dependent robotic manipulation. The harness supports
within-episode context management and is revised using development-stage
execution experience, with the selected version fixed before testing.

\section{Method}
\label{sec:method}

\subsection{Problem Formulation}
\label{sec:problem_formulation}

We consider history-dependent robotic manipulation tasks specified by a
language instruction ${i}$. Each episode may provide initial
context $c_0$, such as a demonstration or previously observed events. {At a policy query time $t$, the robot receives a sensory
observation $o_t$, and the policy predicts an action chunk $A_t$.} The decision problem is
non-Markovian with respect to the instantaneous observation: similar current
observations can require different actions depending on earlier interactions.
{For example, the correct placement target may depend on the order in
which locations were visited in an earlier demonstration, which cannot be
inferred from the current observation alone.}

We define the interaction history $h_t$ and predicted action chunk $A_t$ as
\begin{equation}
    \begin{aligned}
    h_t&=(c_0,o_0,a_0,\ldots,a_{t-1},o_t), \\
    A_t&=(a_t,\ldots,a_{t+K-1}),
    \end{aligned}
    \label{eq:interaction_history}
\end{equation}
where $K$ is the prediction horizon {and $a_{{j}}\in\mathcal{A}$ is a control action}.
The objective is to complete the task
within a bounded interaction budget using a history-conditioned policy,
\begin{equation}
    A_t\sim\pi(\,\cdot\mid {i},h_t).
    \label{eq:history_conditioned_policy}
\end{equation}
{SERA maintains explicit task context alongside recent
observations to support high-level VLM planning. This context is an
approximate record of task state, updated as new execution evidence arrives.
We organize context updates and command selection within an agent workflow,
connecting information retained across interactions with ongoing planning
and execution, as described in Section~\ref{sec:agentic_harness}.}

\subsection{Agentic Harness for Planning}
\label{sec:agentic_harness}
\label{sec:context_management}

\begin{figure*}[t]
\centering
\includegraphics[width=0.95\linewidth]{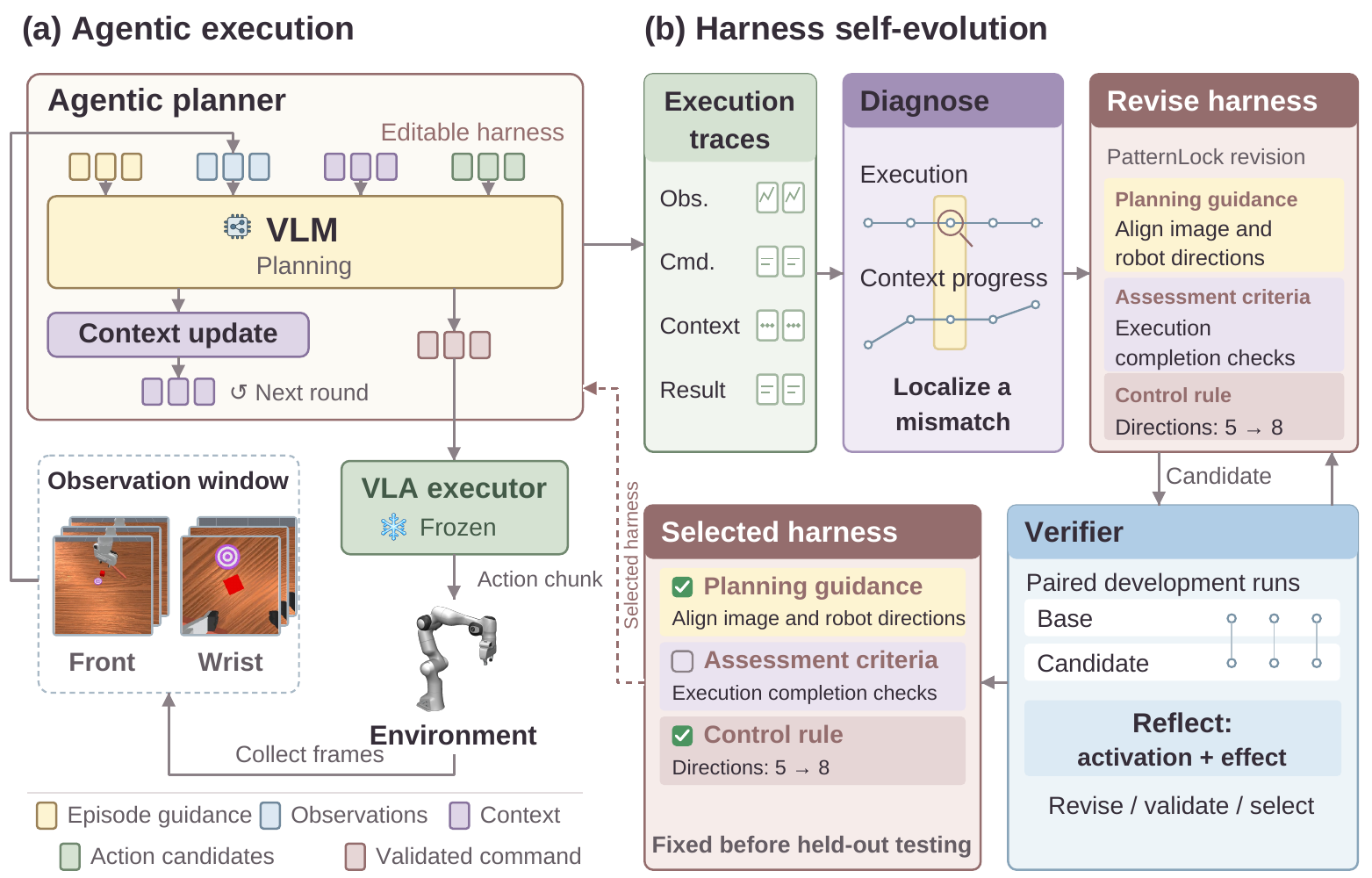}
\caption{\textbf{{Agentic execution and harness self-evolution.}}
\textbf{(a)} The harness maintains task context and uses a VLM to plan commands
for a VLA executor, incorporating execution feedback in subsequent rounds.
\textbf{(b)} Execution traces guide revisions to planning guidance, assessment
criteria, and control rules. A verifier evaluates candidates and informs further
revision and selection; the selected harness is fixed before testing.
Tokens denote information categories and image stacks show observation sequences.}
\label{fig:method_overview}
\end{figure*}

{We adopt an intuitive way to organize a VLM as a task-level
agent: a workflow that maintains context and coordinates interaction with a
fixed VLA executor (Figure~\ref{fig:method_overview}(a)). An editable harness
$H$ supplies task guidance, context, observations, and executor commands,
checks the model's outputs, and manages execution. Self-evolution revises
$H$ using rollout feedback without additional planner training
(Section~\ref{sec:self_evolution}).}

\paragraph{{Task guidance and context.}}
{At initialization, the harness derives episode guidance $g$,
including a task description, from instruction $i$ and initial context $c_0$.
This guidance remains fixed within the episode, while task context $c_n$
changes with new observations and execution feedback. The distinction
preserves the task requirements while allowing the agent's estimate of
execution progress to be updated. The context contains a compact summary
of task progress, decision-relevant facts, execution outcomes, and unresolved
uncertainty, together with control state for tracking lack of progress.
Retaining relevant information after its supporting images leave the recent
observation window makes earlier events available to later planning.
Initialization priors and context fields are detailed in
Appendix~\ref{app:method_context}.}

\paragraph{{Observations and the executor interface.}}
{The workflow provides recent visual observations $V_n$ together
with the previously executed command $u_{n-1}$, so that the VLM can interpret
what happened in relation to what was requested. It also supplies action
candidates $A_{\mathrm{cand},n}$ from the executor's supported command set,
including their required arguments. These candidates provide an explicit
interface for expressing the next decision as an executor command. The
harness checks response structure, command templates, and grounding fields,
and applies configured execution-control rules before dispatch. Candidate
retrieval and command-format details are given in
Appendix~\ref{app:method_interface}.}

\paragraph{{Planning--execution workflow.}}
{At round $n$, the harness assembles $g$, $V_n$, $c_n$,
$u_{n-1}$, and $A_{\mathrm{cand},n}$ for the VLM. Within the same call, the
VLM assesses the effect of the preceding command, updates the task context,
and proposes the next command. Action-specific assessment criteria, when
available, guide attention to execution evidence within this call.
The harness then checks and processes the proposal, stores the resulting
context, and issues the effective command. The combined update is}
\begin{equation}
    ({c_{n+1}},u_n)=\mathcal{P}_{H}({g},V_n,{c_n},u_{n-1};{A_{\mathrm{cand},n}}).
    \label{eq:agentic_update}
\end{equation}
{Here, $\mathcal{P}_{H}$ includes the VLM proposal and the
harness's programmatic checks and execution-control rules. The first round
uses an initial context and no preceding command. Appendix~\ref{app:planning_round}
details the stages and prompt structure.}

{The harness converts $u_n$ into the executor's required input
format. Let $x_t$ denote the executor's native conditioning inputs, including
its observations and any required task or history inputs. At policy query
time $t_n$, the VLA with fixed parameters $\theta$ predicts an action chunk,}
\begin{equation}
    A_{t_n}\sim\pi_{\theta}(\,\cdot\mid x_{t_n},u_n).
    \label{eq:executor_interface}
\end{equation}
{The loop executes a bounded prefix of this chunk over multiple
control steps and collects observations for the next round. Each round
queries the VLA once; the same subtask can continue with newly predicted
chunks. New execution evidence and retained context inform whether to
continue, advance, or recover a missing prerequisite. Harness control state
tracks repeated uncertainty or stagnation and can request reconsideration
or another command selection. The loop ends when the episode terminates or
exhausts its budget, and runtime context resets between episodes
(Appendix~\ref{app:method_context}). {The harness configuration
shapes how this workflow assesses progress and selects commands.}}

\subsection{Execution-Driven Self-Evolution}
\label{sec:self_evolution}

{An initial harness may not anticipate all task requirements,
leading to misunderstandings of the executor and simulation environment,
incorrect judgments of task state from visual observations, or mismatches
between commanded directions and predicted grounding coordinates. Execution-driven self-evolution uses environment rollouts to identify
and revise these limitations. The agent's interpretation of task concepts may
be misaligned with the downstream subtask executor; apparent visual progress
may be mistaken for actual subtask completion; and a command may request motion toward one side while its predicted
grounding point lies on the other. Examining these problems alongside
effective repairs proposed during open-ended model exploration motivates three
categories of harness revisions: \emph{planning guidance} (G),
\emph{assessment criteria} (A), and \emph{control rules} (C).
These categories provide a structured reference for revision.}

{Planning Guidance (G) revisions clarify semantic misunderstandings identified during
self-evolution to guide subsequent planning and recovery. Assessment Criteria (A) revisions address
inadequate visual completion criteria by specifying the observable evidence
the planner should check before judging an execution outcome. Control Rules (C) revisions
address grounding mismatches through programmatic checks that reject spatially
inconsistent outputs before dispatch, alongside rules governing recovery and
task transitions. These revisions modify harness instructions and runtime
control while preserving the executor's command interface.}

{Starting from $H_0$, an analysis agent diagnoses failures from
rollout evidence, a revision agent proposes candidate harness changes, and a
verifier evaluates candidate rollouts to guide further revision and selection
(Figure~\ref{fig:method_overview}(b)). This process updates the harness rather
than model weights. The selected $H^{\star}$ is fixed before held-out testing,
while task context continues to update within each episode. Detailed procedures
and selection rules are given in Appendix~\ref{app:evolution_procedure}.}

\section{Experiments}
\label{sec:experiments}

\begin{table*}[t]
\centering
\caption{\textbf{RoboMME main results.} Task-averaged success rates (\%; higher is better). Memory-augmented policies equip action models with mechanisms to retain and use history; hierarchical policies separate subgoal planning from low-level execution, including systems that also use memory. {SERA} is grouped with hierarchical policies. Baselines use official leaderboard results; {SERA} uses autonomous selection. \textbf{Bold} and \underline{underlined} values denote the best and second-best scores across all listed methods.}
\label{tab:robomme}
\setlength{\tabcolsep}{5pt}
\resizebox{0.95\linewidth}{!}{\begin{tabular*}{\linewidth}{@{\extracolsep{\fill}}lrrrrr@{}}
\toprule
Method & Counting & Permanence & Reference & Imitation & Average $\uparrow$ \\
\midrule
\multicolumn{6}{@{}l}{\textit{VLA baselines}} \\
$\pi_{0.5}$ & 28.78 & 17.00 & 17.17 & 8.78 & 17.93 \\
$\pi_{0.5}$ w/ action history & 29.09 & 22.75 & 14.59 & 11.17 & 19.40 \\
\midrule
\multicolumn{6}{@{}l}{\textit{Memory-augmented policies}} \\
SAM2Act & 35.33 & 26.00 & 16.83 & 7.33 & 21.37 \\
FrameSamp+Modul & \textbf{65.22} & 25.11 & 36.33 & \textbf{51.39} & \underline{44.51} \\
\midrule
\multicolumn{6}{@{}l}{\textit{Hierarchical policies}} \\
{SimpleSG+Gemini} & {39.00} & {13.50} & {22.50} & {18.00} & {23.25} \\
SimpleSG+QwenVL & 44.61 & 19.61 & 25.22 & 26.56 & 29.00 \\
GroundSG+Gemini & 12.75 & 15.75 & 11.25 & 6.50 & 11.56 \\
GroundSG+QwenVL & 38.00 & 39.34 & 31.56 & 21.89 & 32.70 \\
MemER & 48.83 & \underline{53.17} & \underline{38.00} & 29.50 & 42.38 \\
\addlinespace[2pt]
\textbf{{SERA} (Ours)} & \underline{49.00} & \textbf{57.00} & \textbf{69.50} & \underline{30.00} & \textbf{51.38} \\
\bottomrule
\end{tabular*}}
\end{table*}

{We evaluate SERA on history-dependent manipulation tasks in both
simulation and real-world settings. Through task-performance comparisons,
component ablations, and execution cases, we examine how a high-level VLM
operates within an agentic harness and how execution feedback informs
harness revision.}

\subsection{Experimental Setup}
\label{sec:experimental_setup}

\paragraph{{Simulation.}}
{We use RoboMME~\citep{dai2026robomme}, comprising 16 tasks across
Counting, Permanence, Reference, and Imitation. Each SERA configuration is
evaluated on 50 episodes per task (800 total), reporting task-averaged success
rates. Table~\ref{tab:robomme} compares VLA baselines ($\pi_{0.5}$ with and
without action history), memory-augmented policies (SAM2Act and
FrameSamp+Modul), and hierarchical policies (SimpleSG, GroundSG, and
MemER)~\citep{physicalintelligence2025pi05,fang2025sam2act,sridhar2025memer}.
SERA uses GPT-5.6 with \texttt{high} reasoning effort
and RoboMME's released $\pi_{0.5}$ executor fine-tuned for grounded subtasks.
The executor weights remain fixed during self-evolution and evaluation.
Evaluation and implementation details appear in
Appendices~\ref{app:details} and~\ref{app:method_interface}.}

\begin{figure*}[t]
\centering
\includegraphics[width=0.95\linewidth]{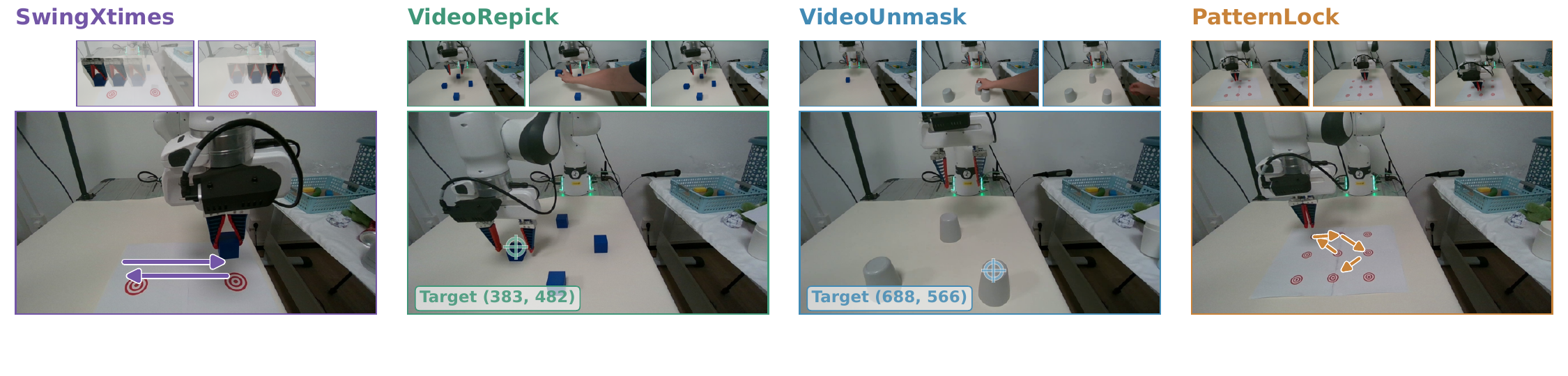}
\caption{{\textbf{Real-world tasks.}
SwingXtimes moves a grasped cube back and forth a
specified number of times. VideoRepick picks the cube selected in a demonstration.
VideoUnmask lifts the cup hiding a cube, tracking its location through covering
and possible cup swaps. PatternLock follows a demonstrated sequence of markers.
Insets show demonstration keyframes or schematic swing motion.
}}
\label{fig:real_world_t8}
\end{figure*}

\paragraph{{Real-world experiments.}}
{We evaluate SwingXtimes, VideoRepick, VideoUnmask, and PatternLock
with 12 trials per task and evaluated configuration, reporting full-task success.
Baselines include $\pi_{0.5}$, conditioned on the task name and current
observations, and Keyframe+GroundSG, which uses Qwen3.5-4B to predict grounded
subgoals from historical keyframes and recent images.
SERA w/o SE uses the initial harness. {SERA applies one round of
harness corrections proposed by GPT-5.6 with \texttt{high} reasoning effort
from failed cases. To reduce the risk of hardware damage, we manually review
the proposals and test only accepted corrections, without iterative rollout
validation. Tasks with no suitable correction retain their initial results.}
Figure~\ref{fig:real_world_t8} illustrates the tasks.
Appendix~\ref{app:real_world_t8} provides implementation and protocol details.}

\subsection{{Main Results}}
\label{sec:main_results}

\paragraph{{Simulation results.}}
Table~\ref{tab:robomme} shows that the complete {SERA} framework achieves
51.38\% overall success, compared with 42.38\% for MemER and 44.51\% for
FrameSamp+Modul. {SERA} obtains the highest scores among the listed methods
on Reference and Permanence, which require using earlier observations
to resolve task-relevant information beyond the current view.
{We also report recorded planner-call and round times from
historical RoboMME evaluations (Table~\ref{tab:runtime}). As in the VLM-based
MemER baseline, planner calls account for most of SERA's recorded round time.}

\begin{figure}[t]
\centering
\includegraphics[width=\linewidth]{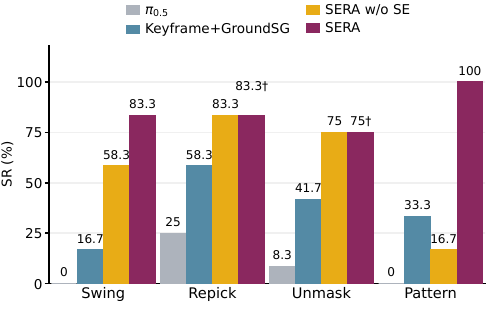}
\caption{{\textbf{Real-world success rates (\%).}
Keyframe+GroundSG uses Qwen3.5-4B.
{SERA reports results after one round of error analysis and human-reviewed corrections.
$\dagger$ marks retained results for tasks where no proposed correction was adopted.}}}
\label{fig:real_world_results_t10}
\end{figure}

\paragraph{{Real-world results.}}
{SERA achieves 85.4\% overall success across four history-dependent
manipulation tasks, compared with 37.5\% for Keyframe+GroundSG and 8.3\%
for $\pi_{0.5}$ ({Figure~\ref{fig:real_world_results_t10}}).
The reported SERA results combine post-correction evaluations on SwingXtimes
and PatternLock with carried-over results on VideoRepick and VideoUnmask,
marked by $\dagger$ in the figure. Error analysis informs corrections to
target-point judgments and relative-position guidance. The final configuration
succeeds in 10/12 trials on SwingXtimes and 12/12 on PatternLock.
Human inspection also identifies cube-grasping failures as an additional
source of failure in SwingXtimes, which remains unresolved.}

\begin{table}[t]
\centering
\small
\caption{{Mean wall-clock latency in historical RoboMME runs (s).
SERA uses the initial harness; MemER is our local reproduction.
{A round includes planner inference, VLA inference, execution of
the action chunk dispatched in that round, and associated overhead.}
{These timings depend on deployment conditions and are provided for reference only.}}}
\label{tab:runtime}
\setlength{\tabcolsep}{3pt}
{\begin{tabular}{lrr}
\toprule
Model & Planner inference (s) & Mean round time (s) \\
\midrule
SERA & 26.60 & 27.33 \\
MemER & 32.31 & 33.97 \\
\bottomrule
\end{tabular}}
\end{table}

\subsection{{Effectiveness of the Harness}}
\label{sec:harness_memory}
\begin{figure}[t]
\centering
\includegraphics[width=\linewidth]{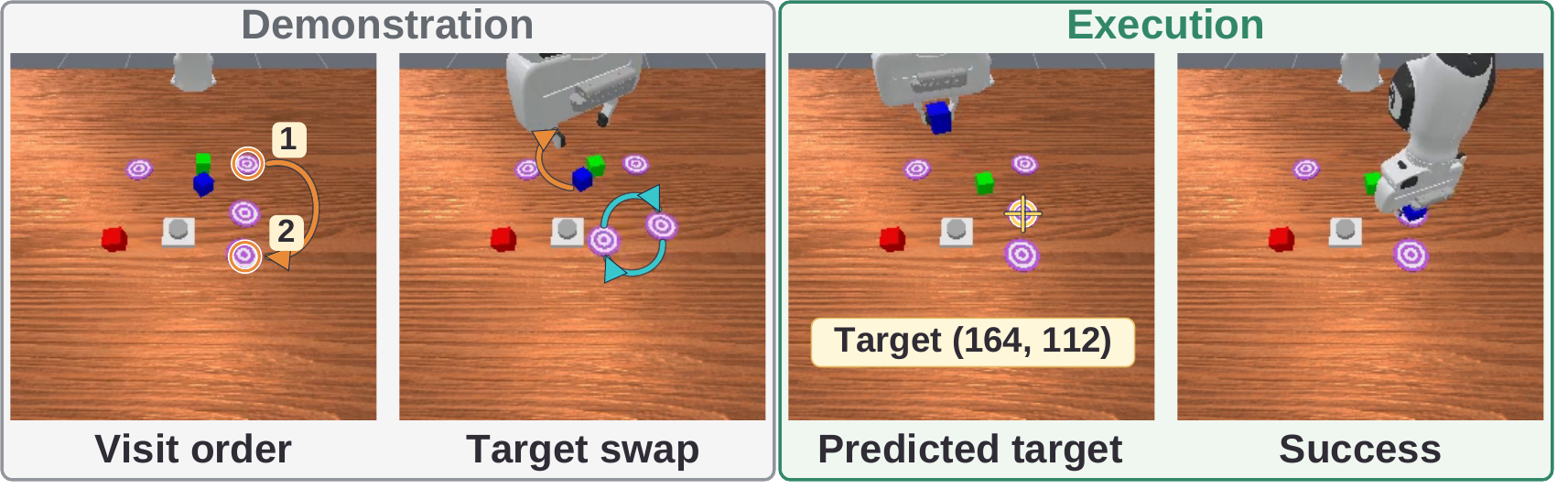}
\caption{{\textbf{Execution process of VideoPlaceOrder.}
The agent analyzes a demonstration to identify the intended target despite
a swap, then uses retained task context to predict its current location and
successfully place the cube. This case shows a test episode.}}
\label{fig:videoplaceorder_t10}
\end{figure}

\begin{table}[t]
\centering
\small
\caption{{Success rates (\%) with matched planner families and sizes. C/P/R/I denote Counting, Permanence, Reference, and Imitation.}}
\label{tab:planner_comparison_t3}
\resizebox{\linewidth}{!}{{\begin{tabular}{llrrrrr}
\toprule
Method & Planner & C & P & R & I & Avg. \\
\midrule
GroundSG & Qwen3.5-4B & 28.00 & 41.00 & 30.00 & 26.50 & 31.38 \\
SERA & Qwen3.5-4B & 39.00 & 51.50 & 53.00 & 13.00 & 39.13 \\
\midrule
GroundSG & Qwen3.8-27B & 49.50 & 38.50 & 49.00 & 28.50 & 41.38 \\
SERA & Qwen3.8-27B & 38.50 & 53.50 & 62.00 & 17.00 & 42.75 \\
\bottomrule
\end{tabular}}}
\end{table}

{To examine whether SERA's performance advantage extends beyond its GPT-based configuration, we compare SERA and GroundSG with Qwen3.5-4B and Qwen3.8-27B, matching planner model families and sizes (Table~\ref{tab:planner_comparison_t3}). SERA achieves average success rates of 39.13\% and 42.75\%, compared with 31.38\% and 41.38\% for GroundSG. The advantage is observed with both Qwen planners, extending the comparison beyond the GPT-based configuration.}

{The most consistent gains occur in Permanence and Reference, where both planner sizes improve. These tasks require information from earlier observations to guide later actions, aligning with the harness's role in maintaining task context throughout execution. GroundSG already uses predicted-subgoal history and grounding information. SERA additionally maintains task-critical facts, such as target identities and operation order, and updates execution progress from recent observations. Together, these allow the planner to use what should be accomplished and what appears to have happened when selecting its next command.}

{SERA with GPT-5.6 also performs particularly well on these two suites in the main results (Section~\ref{sec:main_results}, Table~\ref{tab:robomme}). One possible explanation is that the harness makes the planner's interpretation of visual observations available for later planning by retaining task-relevant information and updating task context with new observations and execution feedback. This may be particularly useful for Permanence and Reference, where earlier observations help resolve task-relevant information beyond the current view.}

{Within the Reference suite, the VideoPlaceOrder execution case (Figure~\ref{fig:videoplaceorder_t10}) illustrates how retained task context supports execution. The agent retains the target identified from the demonstration despite a subsequent swap and keeps that target available across grasp attempts. Once grasping succeeds, the updated execution state supports advancing to placement. The target remains relevant while the next action changes with the observed outcome. This illustrates how the harness connects persistent task information with evolving execution state. The improvements in Permanence and Reference are consistent with the value of this connection.}

{The gains vary across task categories. Counting improves with the 4B planner but declines with the 27B planner, and Imitation declines with both. These experiments compare complete configurations with different fine-tuned weights, initialization, visual inputs, and inference settings, rather than isolating an individual harness mechanism.}

\subsection{{Analysis of Self-Evolution Components}}
\label{sec:self_evolution_results}
\label{sec:ablations}

{An initial harness may leave task concepts misaligned with
executor commands, provide insufficient evidence for completion judgments,
or allow commands with mismatched grounding coordinates. These limitations,
described in Section~\ref{sec:self_evolution}, motivate revisions informed by
execution experience.}

\begin{figure*}[t]
\centering
\includegraphics[width=0.95\linewidth]{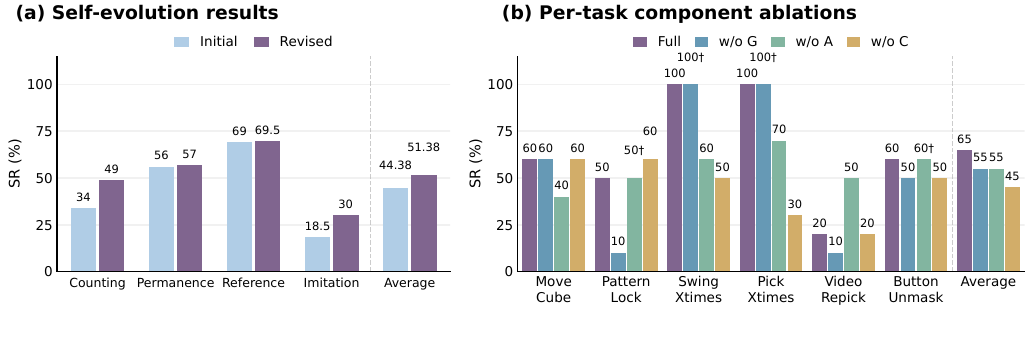}
\caption{{\textbf{Self-evolution results and component ablations.}
(a) Success rates before and after harness revision on 16 tasks with 50 episodes
per task. (b) Per-task success rates (\%) on six tasks with ten episodes each;
Full denotes the complete configuration in this study. Average SR summarizes all six tasks. $\dagger$ marks an unchanged configuration
with no revision of the ablated type, for which Full results are reused rather
than independently rerun. The two panels use distinct evaluation sets.}}
\label{fig:evolution_charts_t10}
\end{figure*}

{Through exploration and rollout feedback in simulation,
self-evolution revises the harness to address two practical questions:
how the agent can better account for the fixed VLA executor's capabilities,
and how it can judge completion using visual observations, remembered context,
and available robot state at runtime. Candidate revisions are evaluated through
further rollouts and selected before held-out testing. The resulting harness
improves overall success from 44.38\% to 51.38\%
({Figure~\ref{fig:evolution_charts_t10}(a)}).
{The initial harness may provide insufficient guidance for
completion judgments and for mapping intended movements to executor commands,
which can hinder execution, especially in Imitation. Self-evolution uses
execution feedback to identify and address these limitations through harness
revisions, potentially contributing to the observed improvements.}
We next examine the roles of the three revision categories through ablation.}

\afterpage{\begin{figure}[t]
\centering
\includegraphics[width=\linewidth]{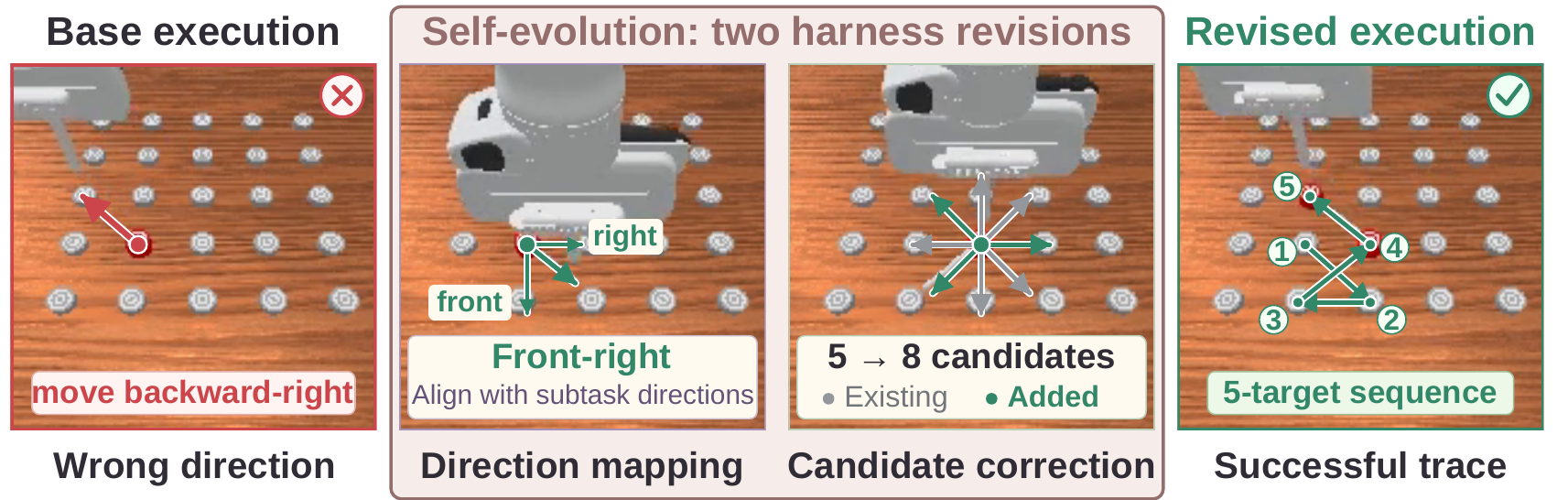}
\caption{{\textbf{Harness repair process in PatternLock.}
Self-evolution diagnoses direction-mapping errors and missing action
candidates, revises the direction guidance and candidate set, and the revised
agent successfully completes the sequence. Direction guides and candidate
colors are schematic, and paths show motion and order.}}
\label{fig:patternlock_t10}
\end{figure}

}

{We conduct a separate component study on six tasks with ten
episodes each, fixing the planner, executor, initialization, test episodes,
and evaluation budgets across configurations. The w/o G and w/o C configurations remove Planning Guidance
and Control Rules revisions, respectively. For w/o A, we replace task-specific
Assessment Criteria refinements with task-independent prompts asking the planner
to recheck whether the previous subtask is complete, while retaining the
controller's required interfaces. All three ablations lower aggregate success
({Figure~\ref{fig:evolution_charts_t10}(b)}), supporting the utility of these revisions in
the tested configurations. Effects vary across tasks, with some ablations
improving success; {Figure~\ref{fig:evolution_charts_t10}(b) reports the
per-task results, with further analysis in Appendix~\ref{app:gac_ablation}.} To understand what these revisions change,
we examine the execution problems they address.}

{In PatternLock ({Figure~\ref{fig:patternlock_t10}}),
examining recorded images alongside the command history and task context
reveals a mismatch between visual directions and executor commands. This
diagnosis motivates a Planning Guidance revision, retained through rollout
evaluation as part of a repair that also expands the available command
candidates (Appendix~\ref{app:patternlock_revision}). The guidance makes the
reference-frame mapping explicit: robot-left corresponds to image-right,
and backward to image-up in the front camera. It thus helps the planner
translate an intended visual target into the appropriate executor command.
With the corrected guidance and candidate set, the agent completes the
illustrated sequence.}

{The other two revision types address completion judgments and
checks on proposed commands. Assessment Criteria revisions specify the observable
evidence needed to judge an execution outcome. In MoveCube, the refinement requires the
peg to move with the gripper before pickup is considered complete. Control
Rules revisions apply programmatic checks before dispatch. In SwingXtimes, the physical
right-side target appears on image-left. A command to that target paired with
an image-right grounding point is therefore rejected before execution.}

\section{Conclusion}
\label{sec:conclusion}

We presented {SERA}, a {self-evolving robotic agent} for history-dependent robotic
manipulation that organizes a high-level VLM within an editable harness around a
fixed VLA executor. The harness manages task context during execution, while
self-evolution uses development-time execution experience to revise planning
guidance, assessment criteria, and control rules. {In simulation, the selected harness is fixed before held-out testing.} On RoboMME, autonomous {SERA} achieves 51.38\%
success, compared with 44.38\% without self-evolution.
{On a real robot, SERA achieves 85.4\% overall
success on history-dependent manipulation tasks.}
{These results motivate further study of agentic planning and
harness adaptation for history-dependent manipulation.}

\paragraph{Limitations and future work.}
{Historical runtime measurements show that planner calls dominate
round time in the recorded deployments.
Repeated calls can therefore limit timely responses in dynamic tasks.} The resulting trade-off between reasoning
quality and execution responsiveness remains to be systematically evaluated.
Future work will investigate more efficient context processing and reasoning
strategies to reduce planning latency while preserving decision quality.

\section*{LLM Usage Statement}
We used OpenAI Codex to assist with literature lookup, code writing,
experiment organization, manuscript drafting and language editing, LaTeX
editing, figure and table layout, PDF compilation, and reference checking.
Its role was assistive, with outputs reviewed and revised by the authors.
The use of language models within the proposed method and experiments is
described in the corresponding sections. The authors made the final research
and editorial decisions and take responsibility for the submitted content.

{
\small
\bibliographystyle{ieeenat_fullname}
\bibliography{references}
}

\clearpage
\appendix
\section{Additional Evaluation Details}
\label{app:details}

\subsection{Metric Aggregation and Evaluation Scope}
\label{app:aggregation}

{For the main comparison,} let $N=16$ be the number of tasks and $n_i=50$ the number of
 evaluation episodes for task $i$. Let $b_{ij}\in\{0,1\}$ indicate
full task success for episode $j$ of task $i$. We report
\begin{equation}
    \mathrm{SR}=\frac{100}{N}\sum_{i=1}^{N}
        \frac{1}{n_i}\sum_{j=1}^{n_i}b_{ij}.
    \label{eq:eval_aggregation}
\end{equation}
Suite-level scores use the same rule restricted to the four tasks in that
suite. Since each task has the same number of episodes, task averaging and
episode pooling coincide. Scores are computed before rounding for display.

On RoboMME, autonomous {SERA} selects intervention bundles before test
evaluation. Candidates can be selected after passing development and
fresh-validation checks. If the search budget is exhausted, a candidate
with one additional success over the baseline and no observed regression
on the development panel can be retained as a fallback. Tasks without a
selected intervention retain the base result. Selected candidates remain
in the aggregate even when their test score decreases; they are not replaced
after observing test outcomes. The comparison therefore reflects the
selection procedure as applied, rather than a per-task maximum over
base and adapted test results.

\subsection{RoboMME Per-Task Results}
\label{app:robomme_tasks}

Table~\ref{tab:robomme_tasks} compares our local reproduction of
MemER~\citep{sridhar2025memer} with the final {SERA} configuration.
Each method is evaluated on 50 episodes per task. The MemER counts come
from a single local evaluation of the released checkpoints; Table~\ref{tab:robomme}
uses the public leaderboard results.

\begin{table*}[htbp]
\centering
\caption{\textbf{Per-task RoboMME results.} Successful episodes out of 50 per task. MemER is our local reproduction using released checkpoints, distinct from the public leaderboard results in Table~\ref{tab:robomme}. {SERA} uses autonomous harness selection. Bold denotes the higher count in each row, including ties.}
\label{tab:robomme_tasks}
\small
\begin{tabular*}{\linewidth}{@{\extracolsep{\fill}}llrr@{}}
\toprule
Suite & Task & MemER (reproduced) & {SERA} \\
\midrule
Counting & BinFill & \textbf{30/50} & 24/50 \\
Counting & PickXtimes & \textbf{41/50} & 35/50 \\
Counting & SwingXtimes & \textbf{36/50} & \textbf{36/50} \\
Counting & StopCube & 2/50 & \textbf{3/50} \\
\midrule
Permanence & VideoUnmask & 43/50 & \textbf{48/50} \\
Permanence & VideoUnmaskSwap & 18/50 & \textbf{46/50} \\
Permanence & ButtonUnmask & \textbf{41/50} & 20/50 \\
Permanence & ButtonUnmaskSwap & \textbf{12/50} & 0/50 \\
\midrule
Reference & PickHighlight & \textbf{35/50} & 27/50 \\
Reference & VideoRepick & 12/50 & \textbf{17/50} \\
Reference & VideoPlaceButton & 14/50 & \textbf{49/50} \\
Reference & VideoPlaceOrder & 18/50 & \textbf{46/50} \\
\midrule
Imitation & MoveCube & \textbf{45/50} & 30/50 \\
Imitation & InsertPeg & \textbf{3/50} & 0/50 \\
Imitation & PatternLock & 10/50 & \textbf{30/50} \\
Imitation & RouteStick & \textbf{3/50} & 0/50 \\
\midrule
\multicolumn{2}{l}{Overall} & 363/800 & \textbf{411/800} \\
\bottomrule
\end{tabular*}
\end{table*}

\subsection{Component Ablation Details}
\label{app:gac_ablation}

\paragraph{Task selection.}
We selected six tasks to examine the failures that motivate planning guidance (G),
assessment criteria (A), and control rules (C) in Section~\ref{sec:self_evolution}.
Three tasks correspond to the repairs analyzed in Section~\ref{sec:self_evolution_results}.
PatternLock requires translating demonstrated spatial relations into directional
commands the executor can follow. MoveCube requires visual evidence that the
gripper has acquired the tool before the planner advances. SwingXtimes requires
each named target to agree with its grounding coordinates across ordered repetitions.
In PickXtimes, verifying pickup and placement determines when to count a repetition
and advance to the next phase. VideoRepick and ButtonUnmask require the agent
to use earlier observations to select a target and act on it later.
These cases let us examine both how the planner accounts for a fixed executor's
capabilities and how it assesses progress from the evidence available at execution time.
Evaluating each ablation configuration requires environment rollouts with repeated
VLM calls, so we limited the study to these six tasks with ten episodes per task.

{The
Full result (39/60) is distinct from the 16-task main comparison (411/800).
The planner, executor, initialization, test episodes, and evaluation budgets
are held fixed across the ablation configurations. Removing G or C disables
the corresponding revisions; w/o A replaces task-specific assessment
refinements with task-independent prompts asking the planner to recheck whether
the previous subtask is complete, and retains the controller's required
interfaces. These interventions do not remove all planning,
assessment, or control functionality. For tasks without a revision of the
ablated type, the configuration is unchanged and the Full result is reused,
as marked in {Figure~\ref{fig:evolution_charts_t10}(b)}.}

{The w/o G, w/o A, and w/o C configurations reduce aggregate success
from 65\% to 55\%, 55\%, and 45\%, respectively. Task-level changes illustrate their roles: removing G reduces
PatternLock from 5/10 to 1/10, replacing A reduces MoveCube from 6/10 to 4/10,
and removing C reduces SwingXtimes from 10/10 to 5/10. Selected paired rollouts
provide concrete illustrations: without G, a PatternLock planner identified
the next target but selected a direction inconsistent with the
front-view-to-executor mapping; without task-specific A, MoveCube advanced
on weaker pickup evidence. In SwingXtimes without C, a wrong-side command
was diagnosed only after execution, illustrating the distinction between
assessment and pre-dispatch constraints. These trajectories illustrate
possible mechanisms rather than establish the cause of each failure.}

{These results support the utility of all three revision types
within the tested configurations. Effects vary across tasks: w/o A improves
VideoRepick from 2/10 to 5/10, and w/o C improves PatternLock from 5/10 to 6/10.
This small study therefore does not establish universal, independent, or
additive benefits, nor that one category is consistently more important.}

\subsection{PatternLock Revision History}
\label{app:patternlock_revision}
{The revision history also shows how verifier feedback guides subsequent changes. On the same six-episode revision panel, the initial harness succeeds in two episodes. A first revision targeting direction interpretation succeeds in three, but inspection of a remaining failure reveals that a required direction is absent from the retrieved candidates. The second revision retains the direction guidance and expands the planner's menu from five retrieved candidates to all eight directional commands already supported by the executor. With this revised candidate set and the direction guidance, the harness succeeds in five of the six episodes. These are pre-test revision results, illustrating how an incomplete repair can be refined through rollout feedback; they do not constitute an isolated comparison of the two repair components.}

\subsection{Runtime Analysis}
\label{app:runtime_analysis}

{We analyze recorded RoboMME evaluations of the initial SERA harness
and our local reproduction of MemER~\citep{sridhar2025memer}. Each configuration
covers 800 episodes, including successes, failures, and timeouts. The SERA
measurements precede self-evolution and do not measure the final evolved
harness. The MemER measurements come from our local deployment, whereas its
main-table success rates are externally reported. Table~\ref{tab:runtime}
summarizes wall-clock latency in these historical runs.}

\paragraph{{Measurement and aggregation.}}
{We average over all recorded planning--execution rounds, comprising
21,224 rounds for the initial SERA harness and 17,769 for MemER. Each round
receives equal weight. This differs from first computing a mean for each
episode and then averaging those means. A round includes the planner call,
one VLA inference call, execution of the resulting action chunk, and other
within-round overhead. Consequently, total round time is not simply the sum
of the two inference-call times. Independent video initialization is excluded
from these round-level measurements.}

{For SERA, planner latency is measured at the client around the
hosted planner request. For MemER, it is measured around the local predictor
call. VLA latency is measured around the client inference call in both
configurations. These measurements include service and communication overhead
where applicable, rather than isolating GPU computation. The 95th-percentile
planner-call latencies are 43.62\,s for SERA and 47.01\,s for MemER, computed
across the recorded calls. Planner calls account for 97.3\% and 95.1\% of total
recorded round time, respectively, showing that planner calls dominate latency
in these deployments.}

\paragraph{{Deployment conditions.}}
{The initial SERA harness uses the hosted
{GPT-5.6 planner with \texttt{high} reasoning effort.}
Its historical evaluation configuration uses eight GPUs and 32 simulator
pipelines for local execution. Our MemER reproduction uses NVIDIA A800 GPUs
with 80\,GB memory. The recorded deployment configuration has 12 simulator
pipelines per four-GPU node. Thus, the MemER timings reflect a concurrent
local deployment rather than an isolated single-GPU benchmark. The two systems
differ in hardware, service access, and concurrency. Network conditions,
request retries, and resource contention can affect the observed call times.}

\paragraph{{Planning frequency and interpretation.}}
{In both measured implementations, each round invokes the planner
once and the VLA once, then executes up to 16 action steps before replanning.
The final round can contain fewer steps. The 16-step interval therefore
corresponds to one VLA call, not two calls producing eight steps each. This
statement concerns the two local implementations used for these measurements.}

{Planning frequency affects the overhead incurred over an episode.
A shorter execution interval requires more planner calls for a given number
of action steps. A longer interval can reduce that call count but delays the
next opportunity to incorporate observations and revise the command. We have
not evaluated this trade-off across planning intervals, so the reported times
should be interpreted together with the 16-step scheduling configuration.}

{Episode budgets also differ in these historical runs. SERA uses
task-specific limits of 300--1,300 action steps, whereas the recorded MemER
protocol specifies a 1,300-step limit. The limit is not retained in every
archived MemER episode summary. Together with the deployment differences,
these conditions prevent a controlled comparison of intrinsic model speed.
The table reports measured operating costs rather than a speed ranking, and
its per-round means are not parallel evaluation throughput. We have not
measured standalone $\pi_{0.5}$ throughput or aggregated API monetary costs.}

\section{{Real-World Experimental Details}}
\label{app:real_world_t8}
\subsection{{Platform and Method Configurations}}
{We use a Franka robot arm for tabletop manipulation and Intel
RealSense cameras for visual observations. The four tasks test memory for
action counts, object identity, the location of an occluded object, and a
demonstrated path (Figure~\ref{fig:real_world_t8}). The evaluation uses 12
trials per task, for 48 trials per configuration evaluated on all four tasks.}

\paragraph{{VLA training and inputs.}}
{The supplied v2 training audit records four $\pi_{0.5}$ runs:
baseline and grounded policies, each with natural or gripper-transition
resampling (probability 0.25). All start from official \texttt{pi05\_base}
weights and complete 50,000 steps with batch size 64 and seed 42 on eight GPUs.
The dataset contains 41,458 windows from 106 recordings processed with three
offsets; these offsets are not independent recordings. The data are recorded
as all-train, without synthetic terminal-hold segments. Baseline inputs are
the task name, current front and wrist images, and an eight-dimensional
state, without historical images, actions, or textual memory. Baseline and
grounded policies are trained separately and have different weights.}

\paragraph{{Keyframe+GroundSG planner.}}
{The v2 planner uses Qwen3.5-4B with LoRA fine-tuning on 5,332
samples for four epochs (rank 16, alpha 32, effective batch 64, learning rate
$10^{-4}$; vision modules frozen). Inputs include the task instruction,
historical keyframes, a recent front-view image window (up to eight images,
one on the first request), and demonstrations where available. Wrist images,
robot states, numerical action histories, and past-subtask text lists are not
planner inputs. Training keyframes follow annotated subtask boundaries and
precede the current window. The output JSON contains
\texttt{current\_subtask}, a grounded natural-language subgoal, and
\texttt{keyframe\_positions}, one-based indices of images to retain.
Following the keyframe-memory approach of MemER~\citep{sridhar2025memer},
this baseline uses visual keyframes as historical context. We use the functional
name Keyframe+GroundSG: Keyframe denotes this visual memory and GroundSG
denotes grounded-subgoal prediction for the low-level executor.}

\paragraph{{Execution interface.}}
{The documented adapter rescales coordinates from 0--1000 to the
256-pixel image coordinate system and passes the subgoal to the grounded
$\pi_{0.5}$ policy in its expected $\langle y,x\rangle$ format. The policy
also receives current front/wrist images, state, and prompt, and predicts
$20\times8$ action chunks: seven absolute joint targets and a gripper command
($-1$ closed, $+1$ open). Keyframe indices serve planner memory management,
not low-level action conditioning. SERA uses the same agent implementation as in the simulation experiments.}

\subsection{{Task Design and Evaluation Protocol}}
\paragraph{{SwingXtimes.}}
{The robot grasps an object and moves it between two target regions.
One cycle counts when the grasped object reaches both regions in the instructed order.
To succeed, the robot must maintain its grasp and stop after exactly the requested
number of cycles. Stopping early or making an extra swing counts as failure.
The left and right conditions each contribute six trials.}

\paragraph{{VideoRepick.}}
{A reference video shows an object being picked up. The robot must
pick up that same object in the current scene, including when a swap has moved
it to a different position. Success requires a stable grasp that lifts the
correct object clear of the support surface. Each of four positions has one
swap trial and two no-swap trials.}

\paragraph{{VideoUnmask.}}
{A reference video shows the target object and its position before
a container covers it. In swap trials, the containers then exchange positions.
The robot must use the initial observation and subsequent motion to identify
the container hiding the target. A trial succeeds when the robot uncovers the
correct container and exposes the target. Uncovering the wrong container or
failing to expose the target counts as failure. Each of three positions has
two swap trials and two no-swap trials.}

\begin{figure}[htbp]
\centering
\includegraphics[width=\columnwidth]{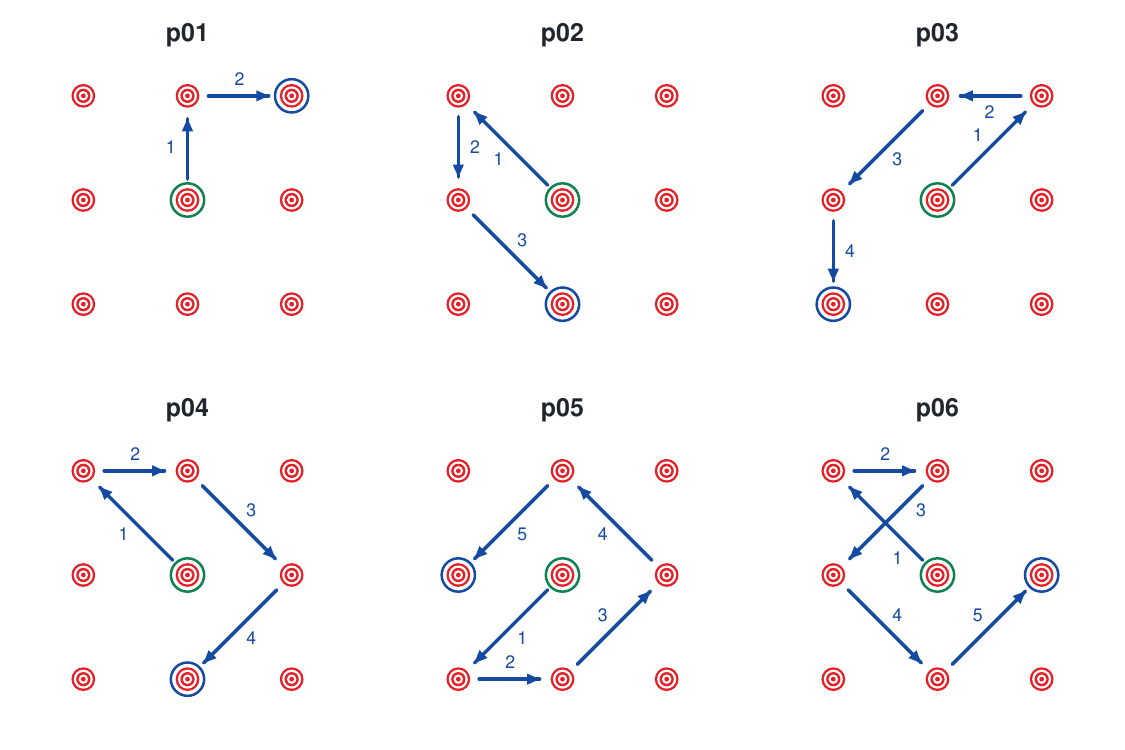}
\caption{Real-world PatternLock paths. Green and blue rings mark the start and end.
Numbered arrows give the order and direction of travel. The evaluation used P03--P06.}
\label{fig:physical_patternlock_paths}
\end{figure}

\paragraph{{PatternLock.}}
{After watching a reference video of motion on a $3\times3$ grid,
the robot traces the demonstrated path with a tracking point on its end effector.
It must start at the demonstrated starting point, visit each target grid point
in order along a continuous path, and come to rest at the final point.
We score the sequence of grid points actually reached. A missed point, a visit
to an incorrect point, or a visit out of order counts as failure. Crossings
between path segments do not count as additional visits. See Figure~\ref{fig:physical_patternlock_paths} for the paths and pp.~\pageref{ar6:first}--\pageref{ar6:second} for photographs of all four tasks. We evaluated P03--P06
three times each. Recording issues precluded evaluation of P01 and P02.}

\subsection{{Reported Outcomes and Provenance}}
{Table~\ref{tab:real_world_counts_t8} gives the success counts
reported by the experimenters for each task and configuration in
{Figure~\ref{fig:real_world_results_t10}}.}
\begin{table}[ht]
\centering\small
\caption{{Real-world task successes (12 trials per cell).
R = VideoRepick, U = VideoUnmask, P = PatternLock, S = SwingXtimes.
{$\dagger$ denotes retained results for tasks where no proposed correction was adopted.}
Keyframe+GroundSG uses Qwen3.5-4B as its high-level model.}}
\label{tab:real_world_counts_t8}
{\begin{tabular}{lrrrr}\hline
Method & S & R & U & P\\\hline
$\pi_{0.5}$ & 0 & 3 & 1 & 0\\
Keyframe+GroundSG & 2 & 7 & 5 & 4\\
SERA w/o SE & 7 & 10 & 9 & 2\\
SERA & 10 & $10^\dagger$ & $9^\dagger$ & 12\\\hline
\end{tabular}}
\end{table}
{The final SERA row combines post-correction results for SwingXtimes
(10/12) and PatternLock (12/12) with the initial results retained for
VideoRepick (10/12) and VideoUnmask (9/12). Together, these outcomes give
41 successes in 48 trials (85.4\%).}

\subsection{{Single-Round Error Analysis}}
{We used one round of error analysis and manual review. GPT-5.6
with \texttt{high} reasoning effort proposed corrections based on failed
cases. We reviewed each proposal before testing it on hardware.
The accepted corrections revised how the planner judged target positions in
SwingXtimes and PatternLock. In PatternLock, they also revised how it
interpreted relative positions along the demonstrated path. We re-tested
these two tasks. For VideoRepick and VideoUnmask, we retained the initial
harness and its results.}

{After correction, successes increased from 7/12 to 10/12 on SwingXtimes
and from 2/12 to 12/12 on PatternLock. Initial experiment notes also record origin-visiting errors and
strict progress judgments in PatternLock, and five stalled SwingXtimes
trials. Post-correction SwingXtimes notes report no recurrence of that stall,
although grasping failures remain. Human inspection identifies failed cube
grasps as an additional failure source. An attempted correction did not
resolve this issue. Limitations of the low-level policy's real-world training
are a possible explanation, but this hypothesis has not been verified.
VideoRepick and VideoUnmask notes report
correct target identification with two and three grasp failures, respectively.
These observations do not isolate the contribution of individual corrections.}

\section{Harness Implementation Details}
\label{app:method_details}

\subsection{RoboMME Command Interface}
\label{app:method_interface}

{The executor is RoboMME's released $\pi_{0.5}$ model fine-tuned
for grounded-subtask execution, using checkpoint
\path{mme_vla_suite/symbolic-grounded-subgoal/79999}.
It receives grounded subtask commands together with visual observations and
robot state. Its weights remain fixed during harness self-evolution and
evaluation. The standalone $\pi_{0.5}$ baseline in Table~\ref{tab:robomme} uses
an externally reported configuration with different conditioning. It should
therefore not be interpreted as an identical-checkpoint comparison.}

The command catalog is extracted from released executor training labels.
Each entry specifies a skill identifier, an exact language template, its
grounding-point count, and recorded task and transition statistics. The base
retrieval scheme first restricts candidates to the known task's entries when
available. It combines rankings for the preceding skill, recorded successors,
and skill frequency using reciprocal-rank fusion, supplying the top five
candidates or all candidates if fewer are available. This scheme does not use
semantic query similarity, even though the retrieval interface accepts text.

The VLM selects a catalog entry and copies its template. Each required
grounding point includes an object description, a source front-camera frame,
and integer coordinates normalized to $[0,999]$ along each image axis. The
harness checks the template, point count, coordinate bounds, and frame validity
before rendering the command. For example, the catalog contains
\texttt{pick up the peg at <point\_1>}; a subsequent hooking template requires
points for the cube and its destination. These are catalog examples, not
measured trajectory outcomes. Zero-point templates require no grounding points.

The VLA retains its native task-goal and observation interface. In the RoboMME
driver, this includes current front/wrist images, proprioceptive state, and
the executor's image/state buffer. This buffer is distinct from the recent
visual window supplied to the high-level planner. The command interface does
not establish physical success; execution evidence is processed in the next
planning round.

\subsection{Context State and Runtime Control}
\label{app:method_context}

{For RoboMME, episode initialization processes the instruction and
initial context using benchmark-provided task-specification priors.
Synchronized front- and wrist-camera observations supply scene-level and
local interaction evidence during execution.}

The RoboMME base planner emits a rewritten progress summary bounded to 1,600
characters. Its prompt requests the current phase, completed steps, task state,
physical state, execution state, and material uncertainty. Structured execution
fields also describe the held object, object support, and gripper state. These
values are model estimates. The description and guidance obtained at episode
initialization remain fixed, while the runtime summary and control state are
updated and reset between episodes.

Execution status distinguishes progressing, completed, uncertain, stagnant,
and regressed outcomes, with a separate initial status. Harness-maintained
counters track repeated lack of progress and stagnation. In the catalog-only
base configuration, they select normal planning, reconsideration of the
current state, or renewed selection among known commands. The counters are
not counts of verified task-completion events. The action-specific visual
check is advisory within the same VLM call; it has no independent Python-side
veto over the visual judgment. Subsequent runtime control determines the
effective command and context stored for the next round.

\subsection{Detailed Self-Evolution Procedure}
\label{app:evolution_procedure}
{
\paragraph{Failure diagnosis.}
An analysis agent examines recorded observations, task context,
issued commands, and outcomes to identify where execution first diverged
from the intended behavior. It considers physical execution and task-progress
reasoning separately, tracing their relationship to the terminal failure.
The diagnosis specifies a repair target, supporting evidence, uncertainty,
and task scope, forming a hypothesis for subsequent rollout evaluation.
Offline diagnostic materials and restrictions on deployable repair triggers
are specified in Appendix~\ref{app:method_selection}.

\paragraph{Candidate harness revisions.}
A revision agent translates the diagnosis into a candidate bundle
with activation conditions and intended behavioral changes. Candidates can
revise three parts of $H$: \emph{planning guidance} directs the VLM's planning and
recovery; \emph{assessment criteria} specify evidence for judging execution
outcomes; and \emph{control rules} govern recovery and task transitions.
Here, planning guidance refers to procedural instructions in the harness,
while $g$ denotes the episode guidance established at initialization.
The harness incorporates candidate changes into prompt construction and runtime
control, allowing the candidate to alter the execution loop in
Section~\ref{sec:agentic_harness}. The revision process operates on these
external components without constructing an additional demonstration dataset
for supervised fine-tuning of the high-level VLM planner.

Candidate revisions are expressed as structured updates to supported harness components. They can change the guidance supplied to the planner or activate registered runtime control mechanisms, while preserving the executor's command interface. This constrains the repair process to changes the execution loop can apply and inspect.

\paragraph{Verification and iterative revision.}
A verifier evaluates candidate harnesses through environment rollouts.
For a given repair target, it compares candidate outcomes with cached
baseline outcomes on the same evaluation episodes, recording repaired
failures, regressed successes, and unchanged outcomes. A subsequent review
examines whether the intervention activated and how it affected execution.
These findings guide another revision, a different repair target, or
rejection within the search budget. Candidates meeting the initial evaluation
criteria can be assessed on additional episodes excluded from their diagnosis
and revision.
A panel is a fixed set of episodes used to evaluate a candidate revision.
Panel construction and selection rules are described in
Appendix~\ref{app:method_selection}.

\paragraph{Selection and deployment.}
The autonomous selection procedure allows both candidates passing
fresh validation and eligible fallbacks retained when the search budget is
exhausted; tasks without a selected revision retain the base harness
(Appendix~\ref{app:method_selection}). The selected configuration $H^{\star}$
is fixed before held-out testing, and test outcomes do not enter its revision
process. During testing, task context $c_n$ continues to update, while
$H^{\star}$ and executor parameters $\theta$ remain fixed.
}

\subsection{Development Selection and Revision Scope}
\label{app:method_selection}

{Offline failure analysis may use benchmark-internal traces or expert
demonstrations in addition to recorded planner and execution data. These are
diagnostic materials: deployed revisions must express their activation
conditions using task identity and runtime-observable information.}

\paragraph{Panel construction.}
For each repair target, we register a revision panel of training and
validation episodes, identified by split and episode ID. We evaluate
successive revisions on the same episodes and compare each outcome with
the cached baseline outcome. When a candidate advances to fresh validation,
we register a second panel from the training and validation splits.
Registration excludes revision-panel episodes, episodes recorded as used
for analysis, and episodes from earlier fresh-validation panels for the
same repair target. For example, the PatternLock revision panel contained
six episodes (four baseline failures and two successes). Its fresh-validation
panel contained four other episodes (three baseline failures and one success).

The autonomous selection protocol permits both candidates passing development
and fresh-validation checks and terminal development fallbacks. In the
reported protocol, a fallback with one additional successful episode and no
observed regression can be retained for selection after search-budget
exhaustion. A task with no selected intervention retains its base result.
The deployed version records the selection route; the aggregate is not
restricted to candidates passing fresh validation. A panel containing only
baseline failures cannot establish retention of previously successful episodes.

Candidate manifests describe task scope, activation conditions, components,
and revision lineage. The checked runtime loads prompt guidance, criterion
overlays, and control rules into their corresponding harness operations. The
manifest format also defines broader extension types, but their presence in
the schema does not establish that they were executed: in particular, the
checked runtime rejects unregistered Python intervention plugins. We do not
treat this interface as evidence of unrestricted online code modification.

\subsection{Planning-Round Operations and Prompt Structure}
\label{app:planning_round}

{A planning round separates input preparation, a VLM proposal,
and programmatic processing of that proposal. Equation~\ref{eq:agentic_update}
summarizes the combined context and command update. The retrieved candidates
$A_{\mathrm{cand},n}$ are supplied as an input to this update.}

\paragraph{{Input preparation by the harness.}}
{The harness assembles episode guidance $g$, recent front- and
wrist-camera observations $V_n$, retained task context $c_n$, the previously
executed command $u_{n-1}$, and runtime control state. It retrieves candidate
commands from the executor's supported catalog and supplies any applicable
action-specific assessment criteria. Retrieval and grounding conventions are
described in Appendix~\ref{app:method_interface}.}

\paragraph{{Proposal by the VLM.}}
{In one call, the VLM assesses the effect of the preceding command,
rewrites the progress summary, and selects a candidate command with grounded
arguments. The supplied assessment criteria guide its interpretation of the
execution evidence within this call. They do not introduce an independent
visual verifier. Execution-state fields are model estimates.}

\paragraph{{Response checks and execution preparation.}}
{The harness checks the response schema, command template,
grounding-point count, coordinate bounds, and source-frame validity. It applies
the configured execution-control rules, stores the resulting context, and
forms the effective command $u_n$. These checks enforce interface requirements
and configured constraints. They do not independently establish physical
success. Equation~\ref{eq:executor_interface} then describes the VLA's generation
of an action chunk from this command and its native inputs. The resulting
execution observations are used in the next planning round.}

\paragraph{{Condensed prompt skeleton.}}
{The following summarizes the documented base planner's prompt
structure. It is not a verbatim prompt or an exhaustive list of
configuration-specific controls.}
\begin{itemize}[leftmargin=*,nosep]
\item {\textbf{Inputs:} fixed task description and episode guidance,
recent images and their frame identifiers, the previous progress summary and
executed command, candidate commands, applicable assessment criteria, and
harness-maintained control state.}
\item {\textbf{Instructions:} assess the previous action from the
available observations, report its status, rewrite the progress summary,
select and copy one candidate command template, ground its required points
in a recent front-camera frame, and return a structured JSON response.}
\item {\raggedright\textbf{Output fields:}
{\footnotesize\sffamily recent\_status, recent\_outcome, refine\_check,
updated\_progress, skill\_key, execution\_state, instruction\_strategy,
current\_\allowbreak subtask\_\allowbreak template, points}.\par}
\end{itemize}
{The progress summary is bounded to 1,600 characters and covers
phase, completed steps, task and physical state, execution state, and
uncertainty. Each grounding point identifies a source frame and integer
coordinates normalized to $[0,999]$.}

\enlargethispage{2\baselineskip}
\paragraph{{Implementation mapping.}}
{In the archived base implementation under
\path{scripts/codex_rollout/}, \path{text_progress_planner.py} assembles model
inputs and handles responses and runtime control. The module
\path{skill_retrieval.py} supplies executor candidates. The \path{prompts/}
directory contains system instructions and the template for round-specific
inputs, and \path{schemas/} specifies the structured response. These locations
identify the components underlying the stages above, rather than a complete
repository listing.}

\clearpage\onecolumn
\section*{Real-world task examples}\label{ar6:first}
\subsection*{SwingXtimes}
The instruction for this example specifies two left--right round trips, starting above the right marker. Each return to the right after visiting the left completes one cycle. After the second return, the robot places the cube on the right marker.
\begin{center}\includegraphics[width=.96\textwidth]{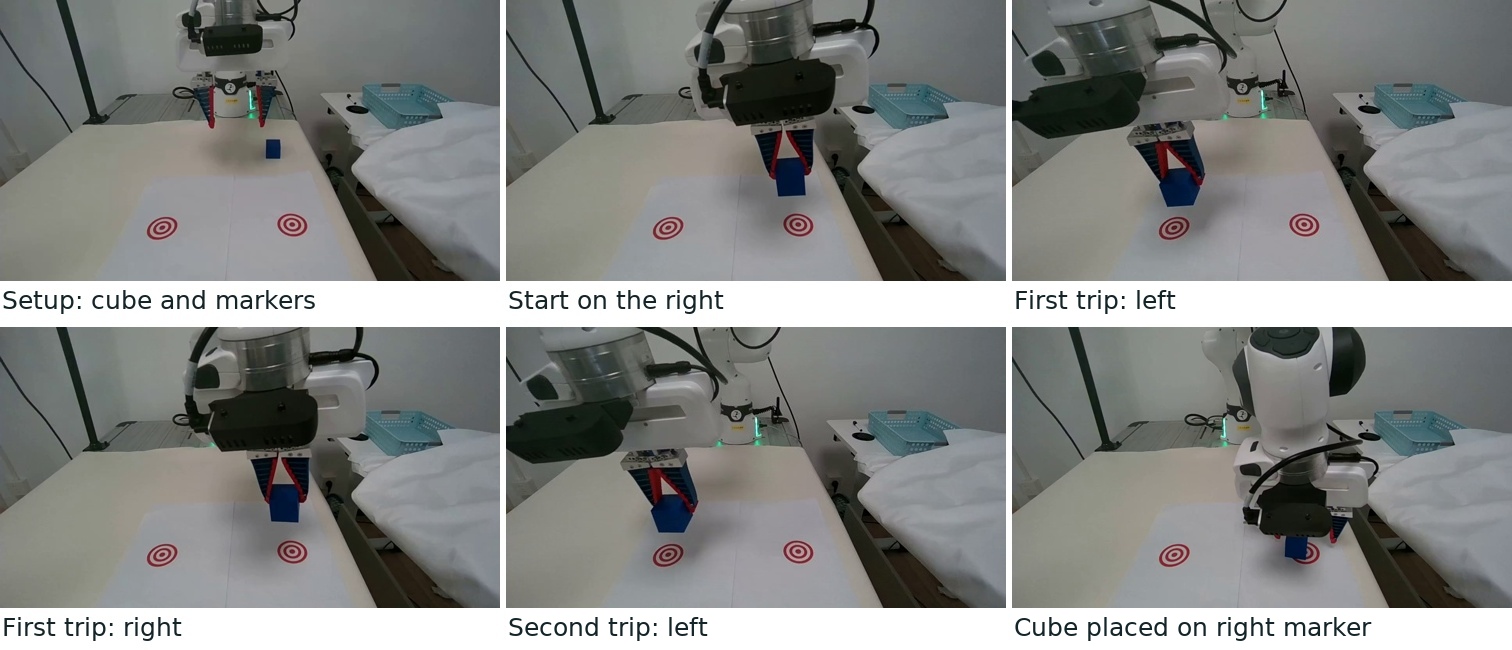}\end{center}
\subsection*{VideoRepick}
The reference identifies one of four similar blue cubes. During the exchange it moves from the back to the right. The upper row shows the reference, and the lower row shows the robot grasping and lifting the target from its new position.
\begin{center}\includegraphics[width=.96\textwidth]{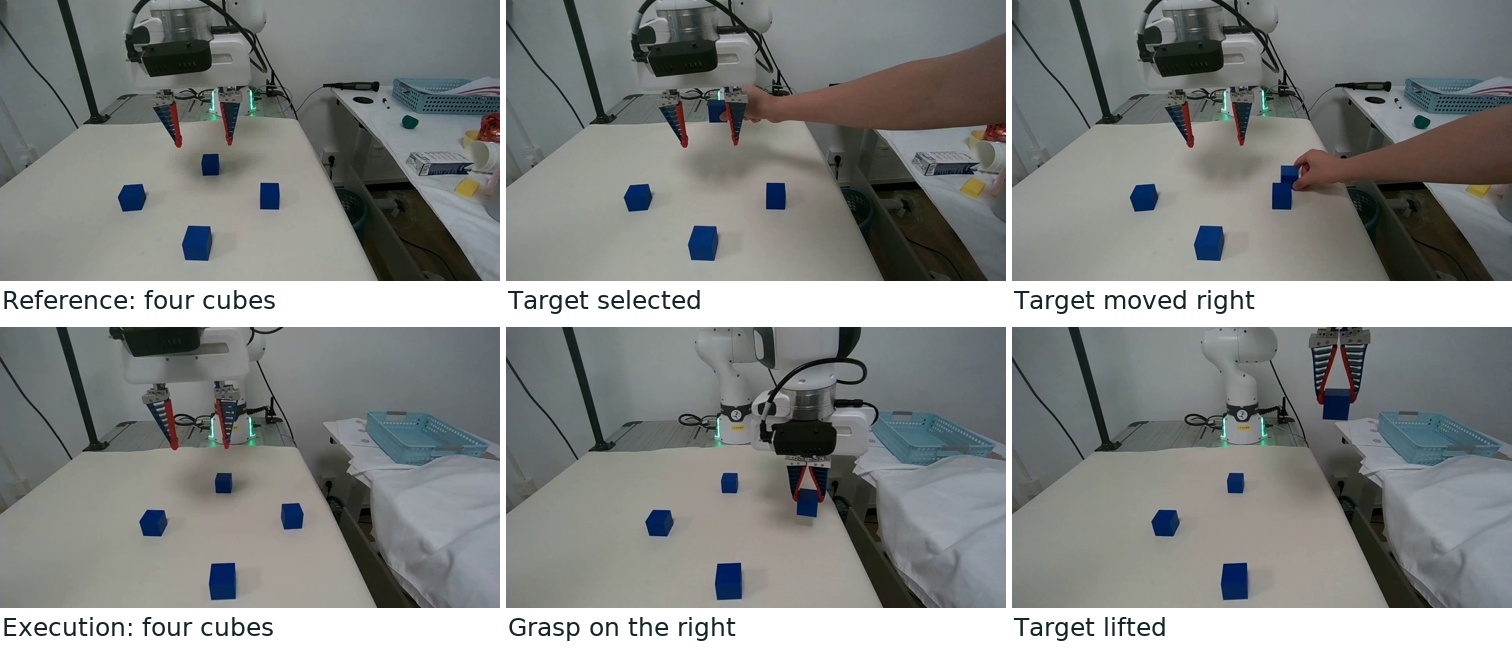}\end{center}
\clearpage
\section*{Real-world task examples (continued)}\label{ar6:second}
\subsection*{VideoUnmask}
The reference shows the blue cube before a cup covers it (upper row). At execution time the cube is hidden among three cups. Raising the selected cup reveals the cube, providing an observable outcome (lower row).
\begin{center}\includegraphics[width=.96\textwidth]{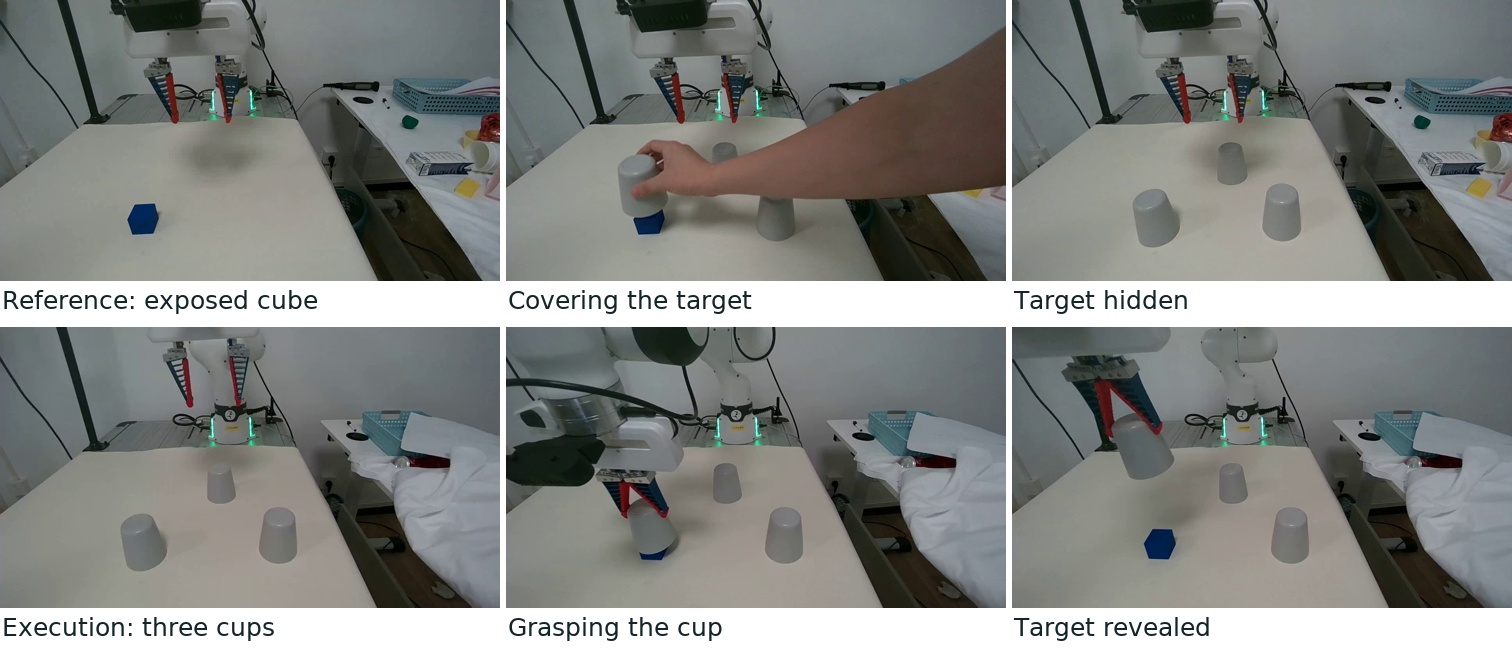}\end{center}
\subsection*{PatternLock}
For path P06, the demonstrated order is $5\rightarrow1\rightarrow2\rightarrow4\rightarrow8\rightarrow6$ on the nine-point board (Fig.~\ref{fig:physical_patternlock_paths}). The photographs show selected poses of the end-effector pointer during execution, labeled with the corresponding waypoints in the demonstrated sequence.
\begin{center}\includegraphics[width=.96\textwidth]{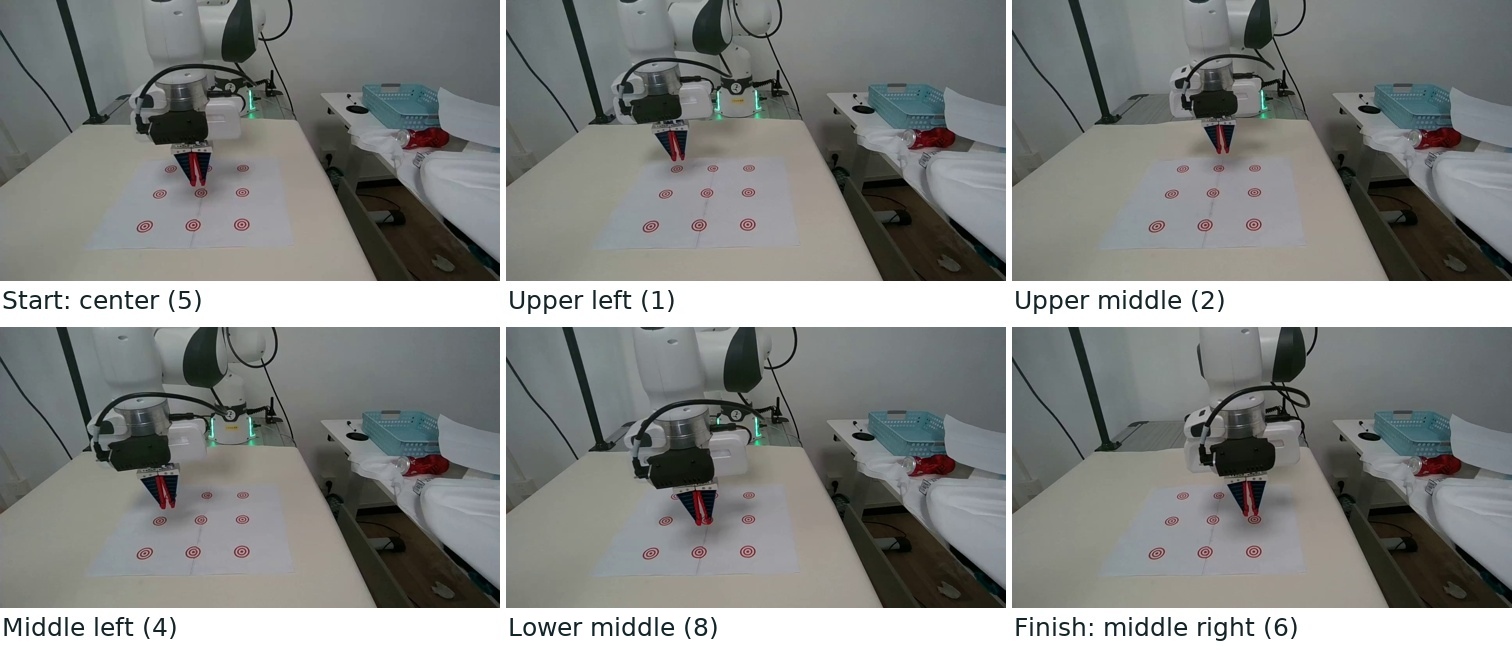}\end{center}

\end{document}